\documentclass[twocolumn, switch]{article} 

\usepackage{preprint}

\usepackage{enumitem}
\usepackage{amsmath, amsthm, amssymb, amsfonts}
\usepackage{bm}
\usepackage{amsmath}
\usepackage{graphicx}
\usepackage{subfigure}
\usepackage[algoruled,algo2e]{algorithm2e}

\usepackage{setspace}
\usepackage{amssymb}
\usepackage{booktabs}
\usepackage{array}
\usepackage{color}
\usepackage{multirow}
\usepackage{multicol}
\usepackage{threeparttable}
\usepackage{url}
\usepackage[page]{appendix}

\usepackage[numbers,square]{natbib}
\usepackage[utf8]{inputenc}	
\usepackage[T1]{fontenc}	
\usepackage{xcolor}		
\usepackage[colorlinks = true,
            linkcolor = blue,
            urlcolor  = blue,
            citecolor = blue,
            anchorcolor = black]{hyperref}	
\usepackage{booktabs} 		
\usepackage{nicefrac}		
\usepackage{microtype}		
\usepackage{lineno}		
\usepackage{float}			

\usepackage{newfloat}
\DeclareFloatingEnvironment[name={Supplementary Figure}]{suppfigure}
\usepackage{sidecap}
\sidecaptionvpos{figure}{c}

\usepackage{titlesec}
\titlespacing\section{0pt}{12pt plus 3pt minus 3pt}{1pt plus 1pt minus 1pt}
\titlespacing\subsection{0pt}{10pt plus 3pt minus 3pt}{1pt plus 1pt minus 1pt}
\titlespacing\subsubsection{0pt}{8pt plus 3pt minus 3pt}{1pt plus 1pt minus 1pt}

\usepackage{graphics}
\usepackage{hyperref}
\usepackage{color}
\usepackage{multirow}
\usepackage{hhline}
\usepackage{subfigure}
\usepackage{lipsum}
\usepackage{pifont}
\usepackage{amsfonts} 
\usepackage{colortbl}

\title{RSFAKE-1M: A Large-Scale Dataset for Detecting Diffusion-Generated Remote Sensing Forgeries
}
\usepackage{authblk}

\author[ ]{Zhihong Tan}
\author[ ]{Jiayi Wang}
\author[ ]{Huiying Shi}
\author[ ]{Binyuan Huang}
\author[ ]{Hongchen Wei}
\author[*]{Zhenzhong Chen}
\affil[ ]{School of Remote Sensing and Information Engineering, Wuhan University}

\begin{document}

\twocolumn[ 
  \begin{@twocolumnfalse} 
  
\maketitle

\begin{abstract}

  Detecting forged remote sensing images is becoming increasingly critical, as such imagery plays a vital role in environmental monitoring, urban planning, and national security. While diffusion models have emerged as the dominant paradigm for image generation, their impact on remote sensing forgery detection remains underexplored. Existing benchmarks primarily target GAN-based forgeries or focus on natural images, limiting progress in this critical domain. To address this gap, we introduce \textbf{RSFAKE-1M}, a large-scale dataset of 500K forged and 500K real remote sensing images. The fake images are generated by ten diffusion models fine-tuned on remote sensing data, covering six generation conditions such as text prompts, structural guidance, and inpainting. This paper presents the construction of RSFAKE-1M along with a comprehensive experimental evaluation using both existing detectors and unified baselines. The results reveal that diffusion-based remote sensing forgeries remain challenging for current methods, and that models trained on RSFAKE-1M exhibit notably improved generalization and robustness. Our findings underscore the importance of RSFAKE-1M as a foundation for developing and evaluating next-generation forgery detection approaches in the remote sensing domain. The dataset and other supplementary materials are available at \url{https://huggingface.co/datasets/TZHSW/RSFAKE/}.

\end{abstract}

\vspace{0.4cm}

  \end{@twocolumnfalse} 
] 

\newcommand\blfootnote[1]{%
\begingroup
\renewcommand\thefootnote{}\footnote{#1}%
\addtocounter{footnote}{-1}%
\endgroup
}

\section{INTRODUCTION}
{\blfootnote{Corresponding author: Zhenzhong Chen, E-mail:zzchen@ieee.org}}

\begin{table*}[t]
\centering
\caption{Comparison with existing remote sensing forgery datasets. \ding{51} indicates the feature is present; \ding{55} indicates absence.}
\label{tab:dataset_comparison}
\resizebox{\linewidth}{!}{%
\begin{tabular}{lcccccccc}
\toprule
\textbf{Dataset} & \textbf{Image Size} & \textbf{Real Vol.} & \textbf{Fake Vol.} & \textbf{Gen. Models} & \textbf{Diffusion?} & \textbf{RS-trained?} & \textbf{Multi-cond.?} & \textbf{FID $\downarrow$} \\
\midrule
FSI~\cite{zhao2021deep}                & 256$\times$256                  & 4,032     & 4,032     & 1 model              & \ding{55} & \ding{51} & \ding{55} & 112.69 \\
Geo-DefakeHop~\cite{chen2024geo}      & 128$\times$128                  & -    & 32,256    & 2 models             & \ding{55} & \ding{51} & \ding{55} & 131.21 \\
DM-AER~\cite{Gupta22dmaer}             & 512$\times$512                  & 119,381        & 999,984        & 1 model              & \ding{55} & \ding{51} & \ding{55} & 45.36 \\
FLDCF~\cite{sui2024fldcf}              & 256$\times$256 / 512$\times$512        & 2,846    & 8,350    & 2 models             & \ding{51} & \ding{55} & \ding{55} & 73.39 \\
\rowcolor{blue!5}
RSFAKE-1M               & 768$\times$768 / 512$\times$512 / 256$\times$256 & 500,000 & 500,000 & 7 methods, 10 models & \ding{51} & \ding{51} & \ding{51} & \textbf{32.26} \\
\bottomrule
\end{tabular}%
}
\end{table*}
In recent years, rapid advances in image synthesis technologies, especially diffusion-based models~\cite{ho2020denoising,rombach2022high}, have made it increasingly easy to generate highly realistic forged images. While substantial research has been conducted on detecting synthetic content in natural and facial imagery~\cite{cai2024AVDeepfake1MLargeScaleLLMDriven,wang2023dire,xia2024advancing}, remote sensing images, which play a critical role in areas such as urban planning~\cite{bai2022urban}, environmental monitoring~\cite{chen2022remote}, agriculture~\cite{ozdogan2010remote}, and national defense~\cite{shimoni2019hypersectral}, have not received sufficient attention in forgery detection research. Forged remote sensing data may mislead downstream analysis, distort decision-making processes, and contribute to the spread of misinformation and public panic. As such, detecting forged remote sensing images has become an increasingly important and urgent challenge.

Detecting forged remote sensing images is a challenging task. The diversity of image content, variation in spatial resolution, and sensor-specific artifacts make it hard for detectors to generalize across scenarios. The emergence of diffusion models further amplifies this difficulty: unlike traditional GANs~\cite{karras2019style,zhu2017unpaired,karras2020analyzing}, diffusion-based generators can produce fine-grained, globally coherent textures that closely match real imagery both visually and statistically~\cite{dhariwal2021diffusion}. Despite their rising prominence, diffusion-generated forgeries remain severely underexplored in the remote sensing community.

The development of forgery detectors typically depends on the availability of suitable datasets. The development of forgery detectors typically depends on the availability of suitable datasets. As shown in Table~\ref{tab:dataset_comparison}, existing remote sensing forgery datasets~\cite{sui2024fldcf,zhao2021deep,chen2024geo,Gupta22dmaer} remain limited in scale and diversity when compared to those in the natural image domain\cite{hong2025wildfake,zhu2023genimage,rahman2023artifact,Wang2022DiffusionDBAL}. Most are focused on GAN-generated images, often relying on a single generation model or containing only a small number of samples. As a result, detectors trained on these datasets struggle to generalize to modern forgeries produced by diffusion models in the remote sensing domain.

To address these limitations, we introduce \textbf{RSFAKE-1M}, a large-scale dataset specifically designed for detecting diffusion-generated remote sensing forgeries. RSFAKE-1M contains 500,000  fake images synthesized by 10 publicly available diffusion models, each trained or fine-tuned on remote sensing data. The dataset supports six types of generation conditions, including text prompts, segmentation masks, edge maps (HED~\cite{xie2015holistically} and Canny~\cite{canny1986computational}), vector maps, and inpainting masks, enabling a broad spectrum of forgery styles. To support balanced and fair evaluation, RSFAKE-1M also includes 500,000 real images sampled from the fMoW dataset, carefully curated to match the resolution and distributional diversity of the fake set.

To validate the utility of RSFAKE-1M, we conduct a comprehensive set of experiments covering six perspectives: the effectiveness of existing detectors, the cross-dataset generalization of unified baselines, intra-dataset generalization across generation types and models, and the impact of training data scale on generalization and robustness. Results show that RSFAKE-1M presents a significant challenge for state-of-the-art detectors, while enabling more reliable and robust model development under both in-domain and cross-domain settings. These findings highlight RSFAKE-1M's value as a strong foundation for advancing research in remote sensing forgery detection.

\section{Related Work}
\subsection{Remote Sensing Image Generation}
Generative models have gained increasing attention in remote sensing, supporting applications such as data augmentation, super-resolution, and scene simulation. Among these, GANs and diffusion models are the two dominant paradigms.
GAN-based approaches laid early foundations for remote sensing image synthesis~\cite{chen2021shuffle,li2022single,xu2022cloud,zhao2022text,wang2019remote,xue2021disaster,wang2023remote}. For example, Chen et al.~\cite{chen2021shuffle} synthesized aircraft images from small datasets using a shuffle-attention GAN, while Li et al.~\cite{li2022single} and Xu et al.~\cite{xu2022cloud} explored super-resolution and cloud removal with attention-enhanced GANs. Despite their effectiveness, GANs often suffer from instability, mode collapse, and limited controllability.
More recently, diffusion models have become a compelling alternative due to their superior fidelity and controllability~\cite{tang2024crs,sastry2024geosynth,khanna2024diffusionsat,espinosa2023generate,immanuel2025tackling,yuan2023efficient,zhang2024rs5m}. GeoSynth~\cite{sastry2024geosynth} generates diverse scenes with textual and geographic conditioning. DiffusionSat~\cite{khanna2024diffusionsat} proposes a foundation model leveraging geospatial metadata for multi-task generation. RSPaint~\cite{immanuel2025tackling} enhances few-shot segmentation via semantic inpainting. With the increasing realism of diffusion-generated imagery, distinguishing real from fake content becomes more difficult, highlighting the urgent need for robust detection benchmarks.

\subsection{Remote Sensing Forgery Detection}

The realism of generative models has raised concerns about geospatial authenticity. Zhao et al.~\cite{zhao2021deep} first introduced the concept of "deep fake geography", using CycleGAN~\cite{zhu2017unpaired} to synthesize geographically false but realistic scenes, and proposed a handcrafted detection method based on spatial, texture, and frequency features.
To overcome limitations of handcrafted approaches, Chen et al.~\cite{chen2024geo} proposed Geo-DefakeHop, a lightweight framework using spatial-frequency features and channel-wise discrimination, achieving robust detection across various manipulations.
While these works focused on detecting GAN-generated images, the challenge of detecting diffusion-based forgeries remains underexplored. Sui et al.~\cite{sui2024fldcf} addressed this by introducing FLDCF, a joint localization-detection framework, along with two benchmark datasets based on diffusion-generated forgeries.
However, existing datasets suffer from limited scale and diversity, often relying on a single model or condition. To address this, we introduce RSFAKE-1M, a large-scale benchmark built with ten diffusion-based remote sensing generation models and six generation conditions, aiming to support comprehensive evaluation of detection generalization and robustness.
\begin{figure*}
\centering
\includegraphics[trim=0pt 9.5pt 0pt 10pt, clip,width=0.9\textwidth]{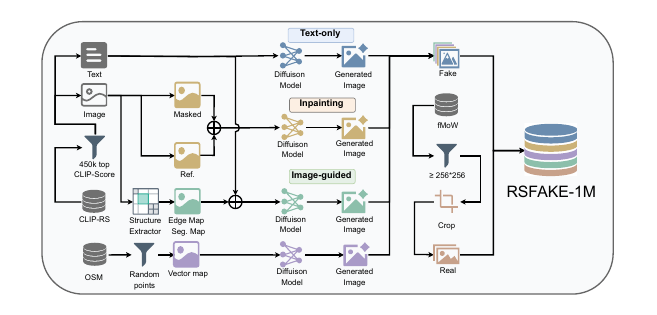}
\caption{Overview of the RSFAKE-1M dataset generation pipeline. The dataset consists of 500,000 fake images generated by 10 diffusion-based models under six different conditions, and 500,000 real images sampled from the fMoW dataset.}
\label{fig:pipeline}
\end{figure*}

\section{RSFAKE-1M Dataset}
\subsection{Fake Image Generation}
\paragraph{\textbf{Generation Model Selection.}} 
RSFAKE-1M focuses on detecting forged remote sensing images that closely mimic real ones in both style and statistics. To ensure domain relevance and high fidelity, we exclusively adopt diffusion-based generation models that have been trained or fine-tuned on remote sensing imagery. Unlike general-purpose diffusion models trained on natural images, these models better capture the characteristics of satellite data. All selected models are publicly available, facilitating reproducibility and responsible research.

Following these principles, we include 10 diffusion-based generators across 7 methods: DiffusionSAT, MapSAT, GeoSynth, GeoRSSD, CRSDiff, SDFRS, and RSPaint. These models vary in resolution, generation condition, and output style. A detailed overview is provided in Table~\ref{tab:generation_models}.
\begin{table}[t]
\centering
\caption{Summary of the 10 diffusion-based generation models used in RSFAKE-1M. All models are trained on remote sensing data and publicly available.}
\label{tab:generation_models}
\small
\begin{tabular}{llll}
\toprule
\textbf{Model} & \textbf{Condition Type} & \textbf{Resolution} \\
\midrule
DiffusionSat-256~\cite{khanna2024diffusionsat}     & Text Prompt                 & 256$\times$256 \\
DiffusionSat-512~\cite{khanna2024diffusionsat}     & Text Prompt                 & 512$\times$512 \\
GeoRSSD~\cite{zhang2024rs5m}                   & Text Prompt                 & 768$\times$768 \\
SDFRS~\cite{yuan2023efficient}                      & Text Prompt                 & 512$\times$512 \\
GeoSynth-text~\cite{sastry2024geosynth}           & Text Prompt                 & 512$\times$512 \\
GeoSynth-sam~\cite{sastry2024geosynth}             & Text + Segmentation Mask    & 512$\times$512 \\
GeoSynth-canny~\cite{sastry2024geosynth}           & Text + Canny Edge Map       & 512$\times$512 \\
CRSDiff~\cite{tang2024crs}                  & Text + HED Edge Map         & 512$\times$512 \\
MapSat~\cite{espinosa2023generate}                  & Vector Map (OpenStreetMap)  & 256$\times$256 \\
RSPaint~\cite{immanuel2025tackling}                  & Masked Image Inpainting     & 512$\times$512 \\
\bottomrule
\end{tabular}
\end{table}
\paragraph{\textbf{Image Generation.}}
To construct a high-quality and diverse dataset of forged remote sensing images, we designed a multi-path image generation pipeline based on diffusion models, as illustrated in Figure~\ref{fig:pipeline}. These paths leverage various generation conditions, including text-only prompts, masked images with reference, and image-guided inputs (e.g., edge maps, segmentation maps, and vector maps).
We began by selecting image-text pairs from the CLIP-RS~\cite{shi2025remotesensingsemanticsegmentation}, a large-scale remote sensing dataset containing 10 million samples, where the captions were refined using GeoChat~\cite{kuckreja2024geochat} to ensure high semantic quality. To maximize textual diversity and relevance, we filtered the dataset using CLIP-Score and retained the top 450,000 image-text pairs as our primary source data.
The text prompts used in the generation process were directly derived from these curated captions. For image-guided generation, edge maps (Canny~\cite{canny1986computational} and HED~\cite{xie2015holistically}), segmentation maps (generated using FastSAM~\cite{zhao2023fast}), and vector maps were extracted from the selected image-text pairs. Correspondingly, the textual descriptions paired with these images were also used as conditions in the generation models to enhance semantic consistency. For vector maps, we followed the MapSat protocol~\cite{espinosa2023generate} and extracted 50,000 samples from the Central Belt region of Scotland using OpenStreetMap~\cite{OpenStreetMap} data. 
For the inpainting setting, we applied 25\% random square masks to the selected images to create masked inputs alongside reference images. Based on the selected diffusion models for remote sensing and the diverse, high-quality generation conditions, we generated 50,000 forged images for each model, resulting in a total of 500,000 high-quality synthetic remote sensing images.

\begin{figure*}[t]
\centering
\includegraphics[trim=0pt 40pt 0pt 5pt, clip,width=\textwidth]{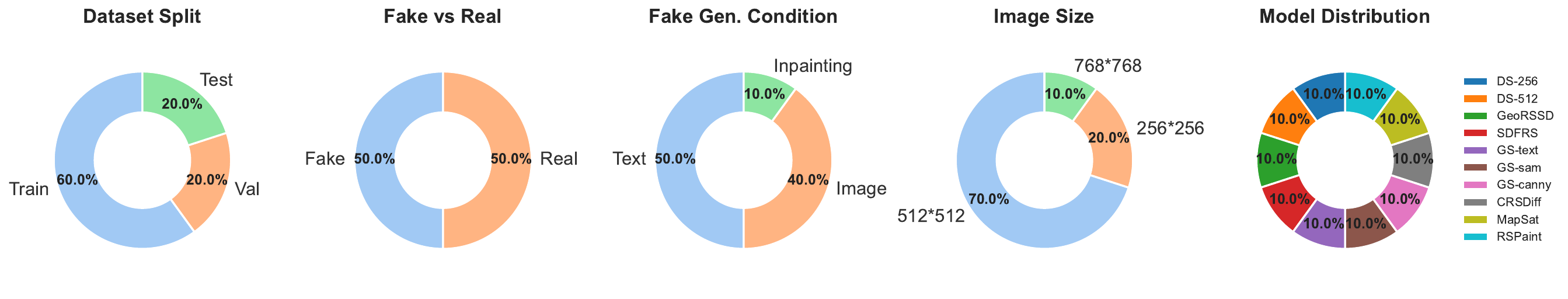}
\caption{Dataset composition of the collected synthetic remote sensing image corpus. From left to right: (a) train/val/test split, (b) proportion of fake and real samples, (c) distribution of fake generation conditions, (d) image size distribution, and (e) sample count per generative model.}
\label{fig:dataset_statics}
\end{figure*}
\begin{table*}[t]
\caption{
Cross-dataset evaluation of existing pretrained forgery detectors. Each model is evaluated on five datasets using its original pretrained weights without any fine-tuning. Bold indicates the best result per column, and \underline{underlined} indicates the second best. For clarity: FLDCF (Fake-LoveDA) and FLDCF (Fake-Vaihingen) denote two versions of FLDCF trained on the Fake-LoveDA and Fake-Vaihingen datasets, respectively; Geo-DefakeHop is trained on the FSI dataset; AIDE (ProGAN) and AIDE (SD v1.4) are trained on ProGAN-generated and Stable Diffusion v1.4-generated natural images, respectively.
}
\label{tab:existing_detectors_results}
\centering
   \resizebox{\textwidth}{!}{%

\begin{tabular}{@{}c|c|ccc|ccc|ccc|ccc|ccc@{}}
\toprule
\multirow{2}{*}{\textbf{Set}}  & \multirow{2}{*}{\textbf{Model}} & \multicolumn{3}{c|}{\textbf{FSI}}                            & \multicolumn{3}{c|}{\textbf{Geo-DeFakeHop}}                  & \multicolumn{3}{c|}{\textbf{DM-AER}}                       & \multicolumn{3}{c|}{\textbf{FLDCF}}                       & \multicolumn{3}{c}{\textbf{RSFAKE-1M}}                       \\ \cmidrule(l){3-17} 
                      &                        & ACC (\%)             & AP (\%)             & AUC (\%)             & ACC (\%)            & AP (\%)            & AUC (\%)            & ACC (\%)           & AP (\%)            & AUC (\%)            & ACC (\%)           & AP (\%)            & AUC (\%)            & ACC (\%)           & AP (\%)             & AUC (\%)            \\ \midrule
\multirow{3}{*}{RS}   & FLDCF (Fake-LoveDA)    & 57.01           & 75.18           & 76.02           & 10.90          & 6.17           & 1.72           & 14.30          & 54.11          & 71.79          & \textbf{85.43} & 78.32          & \textbf{95.72} & 52.33          & 61.33          & 65.75          \\
                      & FLDCF (Fake-Vaihingen) & 59.00           & 56.74           & 58.80           & \textbf{96.43} & 74.60          & 74.75          & 26.42          & 10.92          & 50.94          & {\underline{67.39}}    & 40.49          & {\underline{71.29}}    & 52.56          & 55.45          & 55.83          \\
                      & Geo-DefakeHop (FSI)    & \textbf{100.00} & \textbf{100.00} & \textbf{100.00} & 11.11          & 85.99          & 44.52          & 10.66          & 86.77          & 47.80          & 45.26          & 72.23          & 51.76          & 52.80          & 47.89          & 35.64          \\ \midrule
\multirow{5}{*}{AIGC} & UniversalFakeDetect    & 76.74           & 87.38           & 86.32           & 57.52          & {\underline{97.36}}    & 86.09          & {\underline{53.02}}    & 95.02          & 73.55          & 47.93          & {\underline{79.90}}    & 59.49          & 61.16          & {\underline{66.72}}    & {\underline{67.35}}    \\
                      & AIDE (ProGAN)          & 91.81           & 99.61           & {\underline{99.62}}     & 10.82          & 79.18          & 22.63          & 35.90          & {\underline{95.55}}    & {\underline{74.39}}    & 64.68          & 76.69          & 61.10          & 50.11          & 50.00          & 42.67          \\
                      & AIDE (SD v1.4)         & 61.72           & 76.68           & 73.88           & 39.49          & 99.46          & 96.72          & 16.18          & 90.28          & 51.66          & 43.61          & 69.09          & 41.13          & {\underline{58.85}}    & \textbf{75.65} & \textbf{71.97} \\
                      & SAFE                   & {\underline{99.75}}     & \textbf{100.00} & \textbf{100.00} & 12.06          & 96.28          & 84.24          & 10.80          & 89.79          & 54.02          & 51.76          & 71.95          & 46.17          & 55.10          & 60.47          & 54.16          \\
                      & C2P-CLIP               & 97.15           & {\underline{99.64}}     & 99.55           & {\underline{60.26}}    & \textbf{99.14} & \textbf{94.73} & \textbf{89.44} & \textbf{99.36} & \textbf{95.12} & 46.06          & \textbf{81.82} & 59.24          & \textbf{59.87} & 65.29          & 60.84          \\ \bottomrule
\end{tabular}

}
\end{table*}

\subsection{Real Image Collection}
To ensure that the real and fake images in our dataset are not explicitly paired, we independently collected a separate set of real remote sensing images. In contrast to most existing datasets for remote sensing forgery detection, we maintain an equal number of real and fake samples. This 1:1 ratio helps prevent bias during model training, where class imbalance could otherwise encourage the model to overfit to one label (e.g., always predicting “real” or “fake” to optimize accuracy).
We selected the real images from the fMoW~\cite{fmow2018} (Functional Map of the World) dataset, a large-scale and diverse remote sensing benchmark containing over one million images with varying resolutions and global geographic coverage. The high quality and geographic diversity of fMoW make it well suited for constructing a robust and representative real image set.
Since the images in fMoW vary widely in resolution and aspect ratio, we applied a structured sampling strategy to eliminate distributional biases introduced by size discrepancies between real and fake images. Specifically, we followed these steps:
\begin{enumerate}
  \item Filtered out all images with a minimum side length below 256 pixels;
  \item Calculated the total number of image patches needed to match the resolution distribution of fake images (i.e., 70\% at 512$\times$512, 20\% at 256$\times$256, and 10\% at 768$\times$768);
  \item For each source image, we first extracted a maximally-sized center patch;
  \item For remaining patch slots, we performed additional non-overlapping crops from images that had sufficient remaining area after the initial center crop. The center cropping region was adaptively shifted to ensure that patches do not overlap.
\end{enumerate}

Following this protocol, we obtained a total of 500,000 real remote sensing images, matching the fake image set in both size and resolution distribution.

\begin{table*}[t]
\centering
\caption{Cross-dataset generalization results on five remote sensing forgery datasets. Each model is trained on one dataset (columns) and evaluated on all datasets (rows). Results are measured by accuracy (\%). \textbf{Bold} indicates the best result per column, and \underline{underlined} indicates the second best. All models use ResNet-50 as the backbone, pretrained on ImageNet-1K.}
\label{tab:baseline_crossdataset}

\begin{tabular}{ccccccc}
\toprule
\multirow{2}{*}{\textbf{Train set}} & \multicolumn{5}{c}{\textbf{Test set}} & \multirow{2}{*}{\textbf{Avg.}}\\\cmidrule(lr){2-6}
      & \textbf{FSI}             & \textbf{Geo-DefakeHop}   & \textbf{DM-AER}          & \textbf{FLDCF}           & \textbf{RSFAKE-1M}          &                      \\ \midrule
   FSI           & \textbf{100.00}      & 13.23          & 13.12          & 25.88          & 49.88          & 40.42                 \\
                          Geo-DefakeHop & 50.37          & \textbf{100.00}      & 11.47          & 25.35          & 50.22          & 47.48                  \\
                          DM-AER        & 51.67          & 10.62          & \textbf{100.00}      & 25.43          & 50.46          & 47.64                   \\
                          FLDCF         & {\underline{ 53.97}}    & {\underline{ 91.94}}    & 57.61          & \textbf{87.71} & {\underline{ 66.60}}     & {\underline{ 71.57}}     \\
                          RSFAKE-1M        & 45.72          & 75.80           & {\underline{ 83.05}}    & {\underline{ 60.40}}     & \textbf{99.88} & \textbf{72.97}   \\ \bottomrule

\end{tabular}

\end{table*}
\begin{table*}[t]
\centering
\caption{Cross-subset generalization results across 10 generator-specific subsets within RSFAKE-1M. Each model is trained on one subset (columns) and evaluated on all subsets (rows). Results are measured by accuracy (\%). Diagonal cells indicate performance when training and testing are conducted on the same subset. \textbf{Bold} indicates the best result per column, and \underline{underlined} indicates the second best. All models use ResNet-50 as the backbone, pretrained on ImageNet-1K.}
\label{tab:cross_generator_eval}
\resizebox{\textwidth}{!}{

\begin{tabular}{@{}cccccccccccc@{}}
\toprule
\multirow{2}{*}{\textbf{Train set}} & \multicolumn{10}{c}{\textbf{Test set}} & \multirow{2}{*}{\textbf{Avg.}}\\\cmidrule(lr){2-11}
& \textbf{DS-256} & \textbf{DS-512} & \textbf{GeoRSSD}        & \textbf{SDFRS}         & \textbf{GS-text}   & \textbf{GS-sam}    & \textbf{GS-canny} & \textbf{CRSDiff}        & \textbf{MapSat}         & \textbf{RSPaint}        &             \\ \midrule
DS-256 & \textbf{99.86}   & {\underline{58.03}}      & 51.52          & 56.52          & 63.06           & 61.50           & 59.08          & 52.43          & 71.96          & 57.36          & 63.13          \\
DS-512 & {\underline{72.23}}      & \textbf{99.90}   & 52.14          & 59.09          & 60.22           & 60.82           & 67.53          & 53.10          & 51.68          & 58.09          & 63.48          \\
GeoRSSD          & 51.20            & 52.11            & \textbf{99.96} & 76.44          & 78.52           & 79.74           & 78.84          & 55.33          & 77.08          & 57.47          & 70.67          \\
SDFRS           & 59.70            & 52.98            & 87.80          & \textbf{99.99} & 87.62           & 90.22           & 76.23          & 70.93          & {\underline{91.16}}    & 72.42          & 78.91          \\
GS-text    & 50.01            & 50.02            & 50.20          & 50.32          & {\underline{99.99}}     & {\underline{99.68}}     & 53.59          & 50.00          & 51.22          & 50.10          & 60.51          \\
GS-sam     & 54.44            & 52.93            & 83.02          & 64.31          & \textbf{100.00} & \textbf{100.00} & 78.12          & 51.41          & 69.73          & 56.46          & 71.04          \\
GS-canny   & 69.45            & 56.03            & 98.29          & 79.64          & 99.57           & 99.50           & \textbf{99.97} & 55.97          & 90.53          & 76.27          & {\underline{82.52}}    \\
CRSDiff          & 63.22            & 55.14            & 80.35          & 97.78          & 82.00           & 82.67           & 74.66          & \textbf{99.98} & 88.68          & {\underline{82.17}}    & 80.67          \\
MapSat           & 53.24            & 50.18            & 73.80          & 63.67          & 71.94           & 63.91           & 56.79          & 51.41          & \textbf{99.97} & 55.96          & 64.09          \\
RSPaint          & 67.54            & 57.03            & {\underline{98.94}}    & {\underline{99.62}}    & 98.79           & 98.70           & {\underline{96.99}}    & {\underline{97.19}}    & 99.07          & \textbf{99.60} & \textbf{91.35} \\ \bottomrule
\end{tabular}
}
\end{table*}

\subsection{Dataset Statistics}
RSFAKE-1M contains 1 million remote sensing images: 500,000 real and 500,000 diffusion-generated fake images. The real images are cropped from original fMoW scenes, with each sample inheriting the train/val/test split of its source image, resulting in a 60\%/20\%/20\% division.
Fake images follow the same ratio, randomly split while ensuring each diffusion model contributes equally across subsets. All fake samples are created under one of six generation conditions, including text prompt, segmentation mask, edge map (HED and Canny), vector map and inpainting mask, which are grouped into three categories: text-only, image-guided, and inpainting. Figure~\ref{fig:dataset_statics} illustrates the distribution of samples across different models and generation conditions.

\section{Experiments}
\label{Eval}
In this section, we present a series of comprehensive experiments to assess the proposed RSFAKE-1M dataset in terms of its challenge level, generalization potential, and practical applicability across diverse forgery detection scenarios. 
\subsection{Performance of Existing Detectors}  
We begin by evaluating a range of existing forgery detection models across multiple datasets, including both remote sensing-specific detectors (Geo-DefakeHop~\cite{chen2024geo} and FLDCF~\cite{sui2024fldcf}) and general-purpose AIGC detectors (UniversalFakeDetect~\cite{ojha2023fakedetect}, AIDE~\cite{yan2024sanity}, SAFE~\cite{li2024improving}, and C2P-CLIP~\cite{tan2025c2p}). All detectors are evaluated using their pretrained weights. While certain detectors may have previously been exposed to some of the benchmark datasets, none have been trained on RSFAKE-1M. Despite this potential imbalance, the results offer convincing insights into the generalization limitations of current methods.
For comparison, we adopt four benchmark datasets: FSI~\cite{zhao2021deep}, Geo-DefakeHop~\cite{chen2024geo}, DM-AER~\cite{Gupta22dmaer}, and FLDCF~\cite{sui2024fldcf}. The original data splits are preserved for DM-AER and FLDCF, while for FSI and Geo-DefakeHop, we randomly sample 80\% for training and 20\% for testing. This split protocol is held consistent across all experiments.

The experimental results are summarized in Table~\ref{tab:existing_detectors_results}. Remote sensing-specific detectors generally perform well on the datasets they were trained on, but their performance degrades significantly on out-of-domain datasets. This suggests that current remote sensing detectors suffer from poor cross-domain robustness, potentially due to limited diversity in their training data. In contrast, general-purpose AIGC detectors exhibit relatively stronger generalization capabilities across datasets. However, their performance on RSFAKE-1M remains unsatisfactory. These findings highlight that existing detection methods cannot be directly applied to remote sensing forgeries generated by diffusion models. RSFAKE-1M continues to pose a significant challenge for state-of-the-art approaches. 
Given the architectural and training inconsistencies among existing detectors, we next employ unified baselines to perform controlled cross-dataset generalization analysis.
\subsection{Cross-Dataset Generalization Analysis}     
\label{sec:crossdataset}
To demonstrate the influence of training data on model generalization, we perform cross-dataset training and evaluation. Specifically, we train baseline detectors on the training split of each dataset and evaluate them on the test splits of all available datasets. For the baseline, we adopt a ResNet-50 backbone pretrained on ImageNet-1K, equipped with a lightweight two-layer MLP classifier for binary classification. All models are trained for 15 epochs with a batch size of 64, using the AdamW optimizer with a learning rate of 1e-4 and weight decay of 1e-4. The learning rate follows a cosine annealing schedule with a 5\% warmup period. Data augmentations include resizing, random cropping, and horizontal or vertical flipping. All experiments are conducted on multiple NVIDIA GeForce RTX 4060 Ti GPUs.

The experimental results are summarized in Table~\ref{tab:baseline_crossdataset}. We observe that baseline models achieve strong performance on the test set of their own training dataset, but their accuracy drops notably when evaluated on other datasets. This suggests that the baselines are effective at detecting forgeries within the same data domain, while the true challenge lies in cross-domain generalization. Notably, models trained on RSFAKE-1M generally outperform those trained on other datasets in cross-dataset evaluations, suggesting that RSFAKE-1M contributes to improved model robustness under domain shifts.
\subsection{Intra-Dataset Generalization Analysis}   
Due to its diverse generation sources and condition types, RSFAKE-1M enables a fine-grained evaluation of model generalization within the dataset itself. We consider two orthogonal grouping strategies: (1) by generation condition, dividing RSFAKE-1M into three subsets—text-only, image-guided, and inpainting; and (2) by generative model, dividing the dataset into ten subsets corresponding to the ten diffusion models used. For each subset, we train a baseline model and evaluate its performance across all other subsets to assess intra-dataset generalization. To ensure a fair comparison, all models are trained using identical hyperparameters and training strategies. The same baseline architectures and training configurations described in Section~\ref{sec:crossdataset} are used here.
\begin{figure*}[t]

  \centering
\includegraphics[trim=0pt 8pt 0pt 8pt, clip,width=\textwidth]{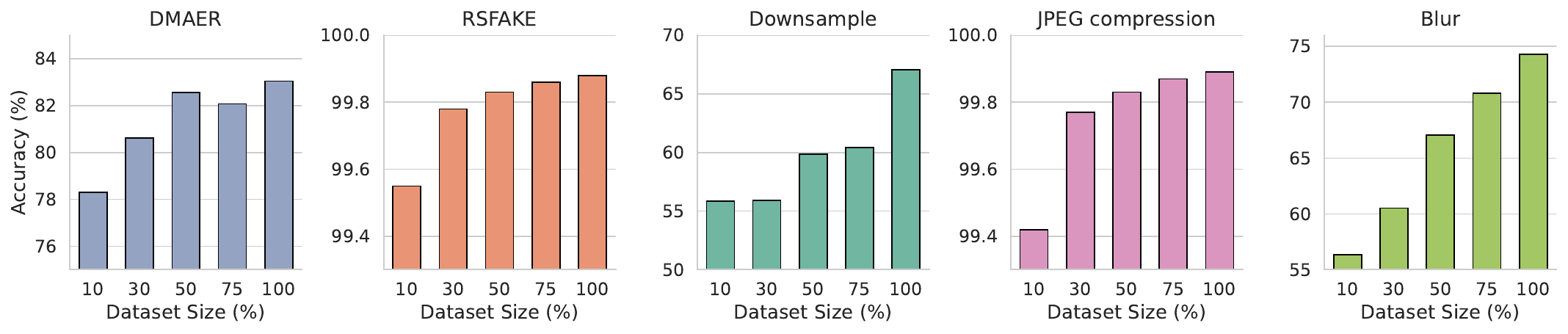}

\caption{Generalization and robustness evaluation of baseline models with ResNet-50 backbone trained on varying portions of RSFAKE-1M. 
The first two plots report accuracy on the RSFAKE-1M and DMAER test sets. 
The remaining three plots show robustness under common image degradations, including downsampling, JPEG compression, and Gaussian blur.}
\label{fig:scale_eval}
\end{figure*}

\paragraph{\textbf{Cross Generator Evaluation.}}
The results of the cross-generator evaluation are presented in Table~\ref{tab:cross_generator_eval}. As expected, models perform best when trained and tested on images generated by the same model. However, performance drops significantly in cross-generator evaluations, indicating that the choice of generative model used for training has a strong influence on detection performance. Moreover, we observe considerable performance differences even among models derived from the same method. For example, the three models based on the GeoSynth approach show notable variation in cross-testing results, suggesting that changes in generation conditions or training data can still lead to distinct forgery patterns, despite architectural similarity.

\paragraph{\textbf{Cross Condition Evaluation.}}
The results of the cross-condition evaluation are presented in Table~\ref{tab:cross_group_eval}. Consistent with earlier observations, models perform better within the same condition they were trained on than across different conditions. Notably, the baseline trained on text-only generated images achieves the highest average performance across all test conditions, suggesting that text-only forgeries provide stronger generalization signals. This may be attributed to the higher diversity inherent in text-only generations, which helps the model learn more robust features of fake imagery. We also observe that models trained on image-guided and inpainting subsets exhibit similar cross-condition behavior, implying a degree of feature similarity between these two types of forgeries.
\begin{table}[t]
\centering
\caption{Cross-condition evaluation results on RSFAKE-1M. Each model is trained on images generated under one condition and tested on all three. Results are reported as classification accuracy (\%). \textbf{Bold} indicates the best result per column, and \underline{underlined} indicates the second best. All models use ResNet-50 as the backbone, pretrained on ImageNet-1K.}
\label{tab:cross_group_eval}
\resizebox{\linewidth}{!}{
\begin{tabular}{@{}ccccc@{}}
\toprule
\multirow{2}{*}{\textbf{Train set}} & \multicolumn{3}{c}{\textbf{Test set}} & \multirow{2}{*}{\textbf{Avg.}}\\
\cmidrule(lr){2-4}
 & \textbf{Text-only} & \textbf{Image-guided} & \textbf{Inpainting} & \\
\midrule
Text-only     & \textbf{99.91} & 95.84          & 85.66          & \textbf{93.80} \\
Image-guided  & \underline{75.81} & \textbf{99.97} & \underline{91.40} & 89.06          \\
Inpainting    & 74.36          & \underline{97.13}          & \textbf{99.60} & \underline{90.36}          \\
\bottomrule
\end{tabular}}
\end{table}
\subsection{Effects of Different Training Scales}           
RSFAKE-1M contains a large number of images, and a key question is whether this scale is truly beneficial for training robust forgery detection models. To validate the necessity and value of such a large-scale dataset, we conduct experiments with varying training set sizes to examine the impact of data scale on model generalization.
 To this end, we construct four subsets containing 10\%, 30\%, 50\%, and 75\% of the full training data, respectively. These subsets are sampled randomly while preserving the balance across all generative models.

For each subset, we train baseline models using the same architectures and training settings as in Section~\ref{sec:crossdataset}. We evaluate the resulting models in two ways: (1) generalization performance, where we assess model performance on both the test set of RSFAKE-1M and the baseline datasets, providing a comprehensive view of in-domain and cross-domain generalization under different training scales; and (2) robustness under degradation, measured by testing on perturbed versions of the RSFAKE-1M test set. Specifically, for the robustness evaluation, we consider three types of common image degradations. For downsampling, each image is resized to 128$\times$128 pixels; JPEG compression is applied with a quality factor of 65; and Gaussian blur is performed with a kernel size of 5 and a standard deviation of 3. These settings are chosen to reflect practical degradations that may occur during image acquisition or transmission.

Figure~\ref{fig:scale_eval} visualizes the generalization and robustness performance of baseline models trained on different-sized subsets of RSFAKE-1M. We observe a consistent upward trend: as the training set size increases, models perform better across both in-domain (RSFAKE-1M test set) and out-of-domain datasets (e.g., DMAER), indicating the importance of scale in enhancing generalization.
In addition to domain generalization, RSFAKE-1M also contributes to improved robustness under degraded conditions. As shown in the right three plots, models trained on larger subsets consistently achieve higher accuracy under downsampling, JPEG compression, and Gaussian blur, demonstrating better resilience to real-world perturbations.
\section{Conclusion}
We introduce RSFAKE-1M, a large-scale benchmark for detecting diffusion-generated remote sensing forgeries. The dataset includes 500K fake images from ten fine-tuned diffusion models under diverse generation conditions, paired with 500K real images. Extensive experiments demonstrate that RSFAKE-1M presents significant challenges for existing detectors, while improving model generalization and robustness when used for training. Its diversity and scale enable fine-grained evaluations across conditions and generation models, making it a valuable resource for advancing research in remote sensing forgery detection.
\bibliography{ms}
\end{document}